\documentclass[runningheads]{llncs}
\usepackage{graphicx}
\usepackage{tikz}
\usepackage{comment}
\usepackage{amsmath,amssymb} % define this before the line numbering.
\usepackage{color}

\usepackage{amsmath}
\usepackage{amssymb}
\usepackage{booktabs}
\usepackage{multirow}
\usepackage{graphicx}
\usepackage{caption}
\usepackage{subcaption}
\usepackage[ruled,linesnumbered]{algorithm2e}
\usepackage{rotating}
\usepackage{optidef}
\newcommand{\avgunder}{\mathop{\mathrm{avg}}\limits}
\newcommand{\norm}[1]{\left\lVert#1\right\rVert}

\usepackage{array, makecell} %
\usepackage[accsupp]{axessibility}  % Improves PDF readability for those with disabilities.

\begin{document}
% \linenumbers
\pagestyle{headings}
\mainmatter
\def\ECCVSubNumber{5769}  % Insert your submission number here

\title{Photo-realistic Neural Domain Randomization} % Replace with your title

\begin{comment}
\titlerunning{ECCV-22 submission ID \ECCVSubNumber} 
\authorrunning{ECCV-22 submission ID \ECCVSubNumber} 
\author{Anonymous ECCV submission}
\institute{Paper ID \ECCVSubNumber}
\end{comment}
%******************

% \begin{comment}
\titlerunning{Photo-realistic Neural Domain Randomization}
% \author{Sergey Zakharov*\inst{1}\orcidlink{0000-0002-6231-6137} \and
% Rareș Ambruș*\inst{1}\orcidlink{0000-0002-3111-3812} \and
% Vitor Guizilini\inst{1}\orcidlink{0000-0002-8715-8307} \and
% Wadim Kehl\inst{2}\orcidlink{0000-0002-2914-8557} \and \\
% Adrien Gaidon\inst{1}\orcidlink{0000-0001-8820-550X} }

\author{Sergey Zakharov*\inst{1} \and
Rareș Ambruș*\inst{1} \and
Vitor Guizilini\inst{1} \and
Wadim Kehl\inst{2} \and \\
Adrien Gaidon\inst{1}}

\authorrunning{S. Zakharov, R. Ambruș et al.}
\institute{Toyota Research Institute, Los Altos, CA \and
Woven Planet, Tokyo, Japan \\
}
% \end{comment}
%******************
\maketitle

\begin{abstract}
Synthetic data is a scalable alternative to manual supervision, but it requires overcoming the sim-to-real domain gap. This discrepancy between virtual and real worlds is addressed by two seemingly opposed approaches: improving the realism of simulation or foregoing realism entirely via domain randomization. In this paper, we show that the recent progress in neural rendering enables a new unified approach we call Photo-realistic Neural Domain Randomization (PNDR). We propose to learn a composition of neural networks that acts as a physics-based ray tracer generating high-quality renderings from scene geometry alone. Our approach is modular, composed of different neural networks for materials, lighting, and rendering, thus enabling randomization of different key image generation components in a differentiable pipeline. 
Once trained, our method can be combined with other methods and used to generate photo-realistic image augmentations online and significantly more efficiently than via traditional ray-tracing. We demonstrate the usefulness of PNDR through two downstream tasks: 6D object detection and monocular depth estimation. Our experiments show that training with PNDR enables generalization to novel scenes and significantly outperforms the state of the art in terms of real-world transfer.

% We apply our approach to the task of 6D object detection, and show that we generalize well to novel scenes and also significantly outperform the state of the art in terms of real-world transfer. 

% Furthermore, our pipeline is fully differentiable, hence the data generation can be optimized adversarially or end-to-end for any downstream task.

\end{abstract}

% terminology: big discussion on twitter about proper terminology: neural fields, neural implicit representations, coordinate-based networks, etc. https://twitter.com/MattNiessner/status/1459563576256090124?s=20  https://twitter.com/vincesitzmann/status/1459606721186381828?s=20
\section{Introduction}

Collecting labelled data for various machine learning tasks is an expensive, error-prone process that does not scale. Instead, simulators hold the promise of unlimited, perfectly annotated data without any human intervention but often introduce a domain gap that affects real-world performance. Effectively using simulated data requires overcoming the \textit{sim-to-real} domain gap which arises due to differences in content or appearance. \textit{Domain adaptation} methods rely on target data (i.e., real-world data) to bridge that gap~\cite{volpi2018adversarial,wang2018high,park2020contrastive,zhang2018multi,zou2019confidence,zheng2021rectifying,liu2021adversarial,guizilini2021geometric}. A separate paradigm that requires no target data is that of \textit{Domain Randomization}~\cite{tobin2017domain,tremblay2018training}, which forgoes expensive, photo-realistic rendering in favor of random scene augmentations. In the context of object detection, CAD models are typically assumed known~\cite{hinterstoisser2019annotation,zakharov2018keep,planche2019seeing} and a subset of lighting, textures, materials, and object poses are randomized. Although typically inefficient, sample efficiency can be improved via differentiable guided augmentations~\cite{zakharov2019deceptionnet}, while content~\cite{kar2019meta,devaranjan2020meta} and appearance~\cite{prakash2021self,mustikovela2021self} gaps can also be addressed by leveraging real data. However, a significant gap remains in terms of the photo-realism of the images generated. As an alternative, recent work~\cite{alghonaim2021benchmarking,guizilini2021geometric} has shown that downstream task performance can be improved by increasing the quality of synthetic data. However, generating high-quality photo-realistic synthetic data is an expensive process that requires access to detailed assets and environments, as well as modeling light sources and materials inside complex graphics pipelines which are typically not differentiable.

We propose a novel method that brings together these two separate paradigms by generating high-quality synthetic data in a domain randomization framework. We combine intermediate geometry buffers (\textit{"G-buffers"}) generated by modern simulators and game engines together with recent advances in neural rendering~\cite{richter2021enhancing,nalbach2017deep,alhaija2018geometric}, and build a neural physics-based ray tracer that models scene materials and light positions for photo-realistic rendering. 
% The materials and lights themselves are represented and sampled from separate neural networks that form the basis of our randomization scheme. 
% Thanks to its geometric input, our ray tracer generalizes to novel scenes and novel object configurations. Although our proposed pipeline is generic in nature, we quantify the usefulness of our synthetic training for the specific task of 6D object detection in a zero-shot setting (i.e., without using any real-world data) on two datasets, and demonstrate that our method presents a distinct improvement over current SoTA approaches.
Our Photo-realistic Neural Domain Randomization (PNDR) pipeline learns to map scene geometry to high quality renderings and is trained on a small amount of high-quality photo-realistic synthetic data generated by a traditional ray-tracing simulator. Thanks to its geometric input, PNDR generalizes to novel scenes and novel object configurations. Once trained, PNDR can be integrated in various downstream task training pipelines and used online to generate photo-realistic augmentations. This alleviates the need to resort to expensive simulators to generate additional high-quality image data when training the downstream task. Our method is more efficient in terms of time (PNDR renderings are generated 3 orders of magnitude faster than traditional simulators), space (PNDR renderings are generated on-the-fly during training and therefore do not need storage space) and leads to better generalization. Although our proposed pipeline is generic in nature, we quantify the usefulness of our synthetic training for the specific tasks of 6D object detection and monocular depth estimation in a zero-shot setting (i.e., without using any real-world data), and demonstrate that our method presents a distinct improvement over current SoTA approaches.

In summary, our contributions are:

\begin{itemize}
    \item We unify photo-realistic rendering and domain randomization for synthetic data generation using neural rendering;
    \item Our learned deferred renderer, \emph{RenderNet}, allows flexible randomization of physical parameters while being $1,600 \times$ faster than comparable ray-tracers;
    \item Our \emph{Photo-realistic Neural Domain Randomization (PNDR)} approach yields state-of-the-art zero-shot sim-to-real transfer for 6D object detection and monocular depth estimation, almost closing the domain gap;
    \item We show that realistic physics-based randomization, especially for lighting, is key for out-of-domain generalization.
\end{itemize}

\section{Related Work}

\noindent\textbf{Domain Adaptation.} Due to the domain gap, models trained on synthetic data suffer performance drops when applied on statistically different unlabelled target datasets. Domain Adaptation is an active area of research~\cite{csurka2017domain} with the aim of minimizing the \textit{sim-to-real} gap. Common approaches rely on adversarial learning for feature or pixel adaptation~\cite{bousmalis2017unsupervised,volpi2018adversarial,ganin2016domain}, paired~\cite{wang2018high} or unpaired~\cite{zhu2017unpaired,park2020contrastive,lee2018diverse} image translation, style transfer~\cite{zhang2018multi}, refining pseudo-labels~\cite{zou2019confidence,zheng2021rectifying,liu2021adversarial}, or unsupervised geometric guidance~\cite{guizilini2021geometric}.

 \begin{figure*}[t]
	\centering
	\includegraphics[width=1\linewidth]{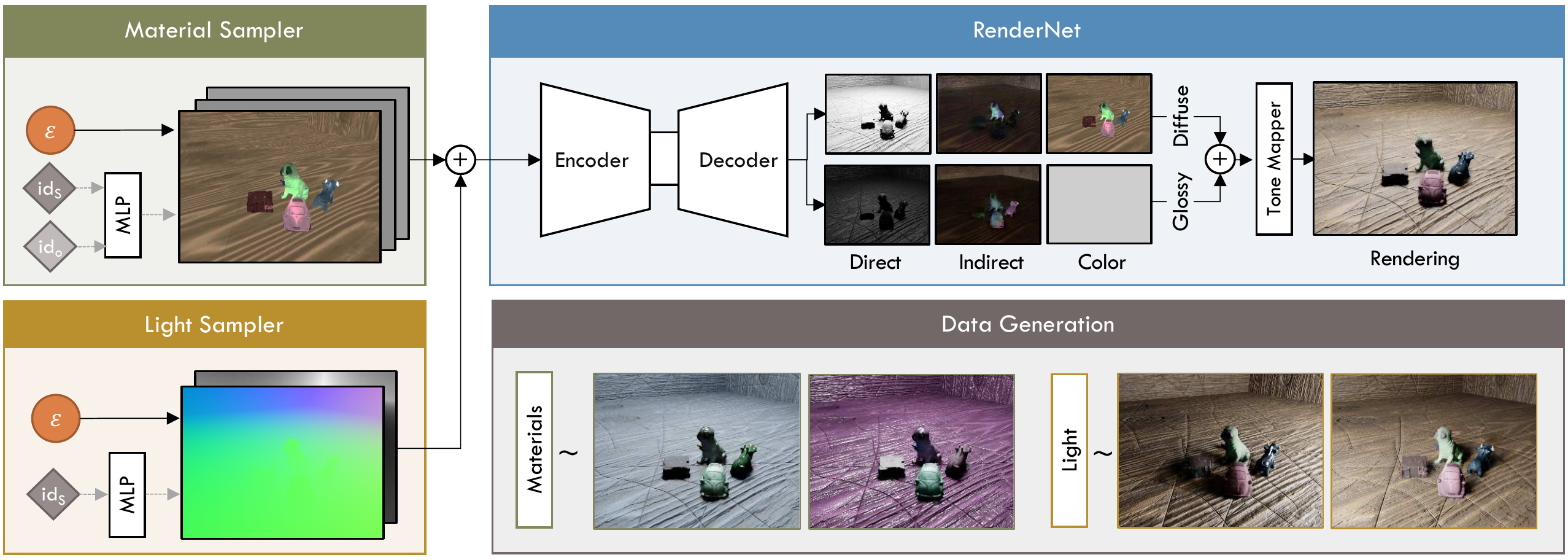}
	\caption{\textbf{PNDR Architecture}. The main component of our domain randomization method is the ray tracer approximator (RenderNet). It takes a G-buffer as well as random material maps and light maps produced by corresponding samplers and generates intermediate light outputs. These outputs are then combined using a tone mapper to generate a final rendering. The lower-right row shows different material and light samples (e.g., roughness, specularity, light position). }
% 	Since PNDR is differentiable, instead of randomly sampling light and material maps we can use the downstream task loss to guide sampling and making the downstream method more robust.
	\label{fig:pndr}  
\end{figure*}

\noindent\textbf{Domain Randomization.} A different approach to closing the sim-to-real gap relies on generating augmentations of the input data through random perturbations of the environment (e.g., lights, materials, background)~\cite{tobin2017domain,tremblay2018training,hinterstoisser2019annotation}. The aim is to learn more discriminative features that generalize to other domains. While simple and inexpensive, this method is sample inefficient because the randomization is essentially unguided with many superfluous (or even harmful) augmentations, and it rarely captures the complexity and distribution of real scenes. Differently, procedurally generating synthetic scenes~\cite{prakash2019structured} can preserve the context of real scenes while minimizing the gaps in content~\cite{kar2019meta,devaranjan2020meta,hodavn2019photorealistic} and appearance~\cite{prakash2021self,zakharov2018keep,planche2019seeing}. While some of these methods require expensive, bespoke simulators~\cite{kar2019meta,devaranjan2020meta}, pixel-based augmentations can be generated differentiably and combined with the task network to generate adversarial augmentations~\cite{zakharov2019deceptionnet}. 
% Our work falls under this paradigm, however, we model traditional graphics rendering pipelines with neural networks, which allows us to generate photo-realistic augmentations. 
Similarly to~\cite{zakharov2019deceptionnet} our pipeline is also differentiable, however while~\cite{zakharov2019deceptionnet} is limited to handcrafted image augmentations where respective parameters are sampled from artificial distributions, our method approximates a material-based ray tracer simulating the physical process of light scattering and global illumination, enabling effects such as shadows and diffuse interreflection. Our augmentations are solely based on light and material changes, thus reducing the randomization set to physically plausible augmentations. Moreover, as opposed to \cite{zakharov2019deceptionnet}, we assume no color information of the objects of interest, making our method more practical for real-world applications.

\noindent\textbf{Photo-Realistic Data Generation.} Although expensive to generate, high-quality synthetic data (i.e., photo-realistic) can increase model generalization capabilities~\cite{alghonaim2021benchmarking,guizilini2021geometric}. The task of view synthesis allows the rendering of novel data given a set of input images~\cite{tewari2021advances}. Neural Radiance Fields~\cite{mildenhall2020nerf} overfit to specific scenes and can generate novel data with very high levels of fidelity, while also accounting for materials and lights~\cite{boss2021nerd,srinivasan2021nerv,boss2021neural}. Alternative methods use point-based differentiable rendering~\cite{ruckert2021adop,aliev2020neural} and can optimize over scene geometry, camera model, and various image formation properties. While these methods overfit to specific scenes, recent self-supervised approaches learn generative models of specific objects~\cite{mustikovela2021self} and can render novel and controllable complex scenes by exploiting compositionality~\cite{niemeyer2021giraffe}. While neural volume rendering and point based techniques can yield impressive results, other methods aim to explicitly model various parts of traditional graphics pipelines ~\cite{richter2021enhancing,nalbach2017deep,alhaija2018geometric,janner2017self,thies2019deferred}. Our work is similar to~\cite{richter2021enhancing} in that we also use intermediate simulation buffers to generate photo-realistic scenes. However, while~\cite{richter2021enhancing} relies on real data and minimizes a perceptual loss in an adversarial framework, we focus on the task of 6D object detection in a zero-shot setting using only object CAD model information and no real images.

% However, while~\cite{richter2021enhancing} relies on real data and minimize a perceptual loss with the aim of increasing synthetic data realism, we only use synthetic data and integrate our differentiable ray-tracer in an end-to-end differentiable pipeline cast as a min-max optimization problem between task loss minimization and adversarial data perturbation.

\noindent\textbf{6D Object Detection.} \emph{Correspondence-based} methods~\cite{zakharov2019dpod,li2019cdpn,jafari2018ipose,hodan2020epos,park2019pix2pose,peng2019pvnet} tend to show superior generalization performance in terms of adapting to different pose distributions. However, they use PnP and RANSAC to estimate poses from correspondences, which makes them non-differentiable. Additionally, they are very reliant on the quality of these correspondences, and errors can result in unreasonable estimates (e.g., behind the camera, or very far away). Conversely, \emph{regression-based} methods~\cite{zhou2019objects,engelmann2021points,labbe2020cosypose} show superior performance for in-domain pose estimation. However they do not generalize very well to out-of-domain settings. To validate our method we implement a correspondence-based object detector, which allows us to also evaluate instance segmentation and object correspondences in addition to the object pose regressed.
% \sz{Do we need a section on depth estimation?}

% CenterNet~\cite{zhou2019objects}, Points2Objects~\cite{engelmann2021points}, CosyPose~\cite{labbe2020cosypose}. However they do not generalize very well to out-of-domain settings. 

% DPOD~\cite{zakharov2019dpod}, CDPN~\cite{li2019cdpn}, iPose~\cite{jafari2018ipose} EPOS~\cite{hodan2020epos}, Pix2Pose~\cite{park2019pix2pose}, PVNet~\cite{peng2019pvnet}.  However, they use PnP and RANSAC to estimate poses from correspondences, which makes them not end-to-end differentiable. Additionally, they are very reliant on the quality of these correspondences, and errors can result in unreasonable estimates (e.g., behind the camera, or very far away). Conversely, \emph{regression-based} methods show superior performance for in-domain pose estimation. CenterNet~\cite{zhou2019objects}, Points2Objects~\cite{engelmann2021points}, CosyPose~\cite{labbe2020cosypose}. However they do not generalize very well to out-of-domain settings. 

% \textbf{Cons:} Not end-to-end, all use PnP+RANSAC to estimate final pose from correspondences. When correspondences are bad can result in an unreasonable pose (behind the camera, miles away). \textbf{Pros:} superior generalization properties in terms of adapting to different pose distributions.
% \textbf{Regression-based}: CenterNet~\cite{zhou2019objects}, Points2Objects~\cite{engelmann2021points}, CosyPose~\cite{labbe2020cosypose}. \textbf{Cons:} Pose distribution learned, bad generalization for out of domain poses. \textbf{Pros:} superior for in-domain pose estimation.
\section{Photo-realistic Neural Domain Randomization}
\label{sec:methodology}

% \subsection{Photo-realistic Domain }

Our photo-realistic neural domain randomization (\textit{PNDR}) approach consists of two main components: a neural ray tracer approximator (RenderNet), and sampling blocks for material and light. To increase perceptual quality and realism, the network outputs are passed through a non-linear tone-mapping function which yields the final rendering. We now describe the main two components of PNDR. All other implementation and training details are provided in the supplementary. 
% \ra{Maybe a prelimiaries section here that describes the geometric scene representations and how that's generated? Added the rebuttal text and fig below, needs refinement}
% \sz{Guidance doesn't work, need to scrap that $\rightarrow$ }Furthermore, we present our domain randomization scheme that leverages PNDR for task-specific training.

\subsection{Geometric Scene Representation}
\label{subsec:geometric_scene_representation}

\begin{figure}[t]
\vspace{-2mm}
	\centering
	\includegraphics[width=1\linewidth]{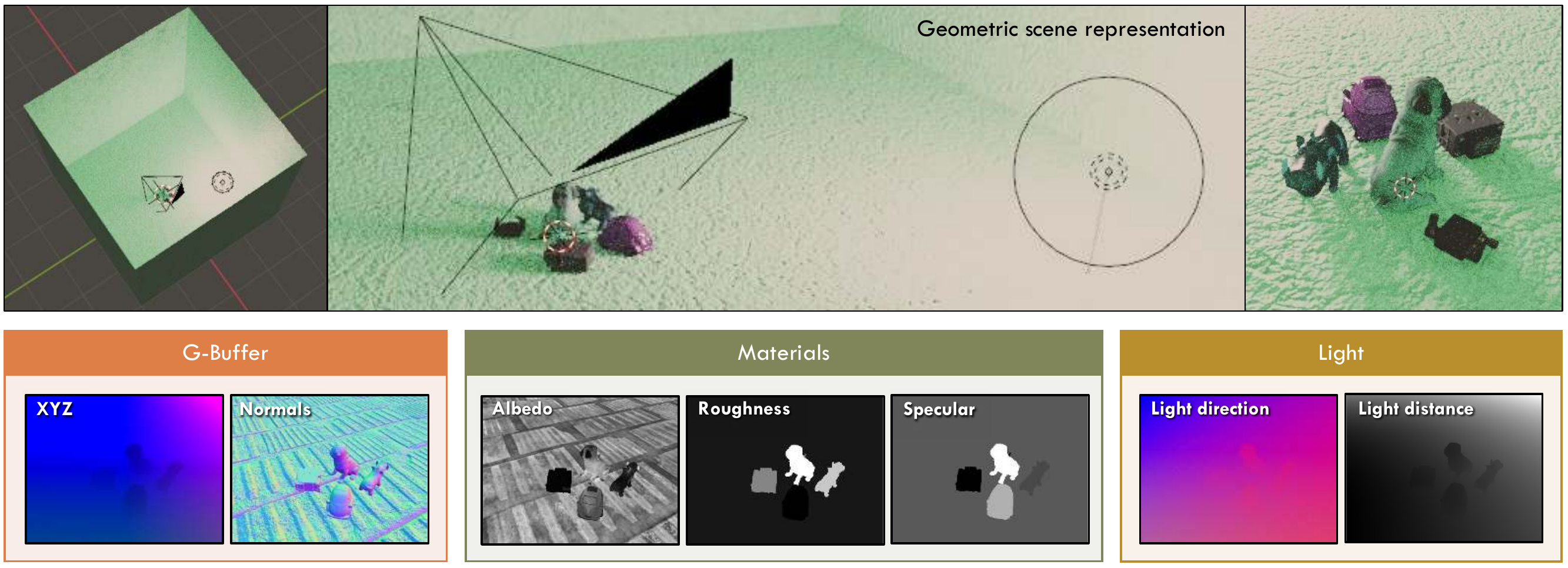}
	\caption{\textbf{Geometric scene representation.} Visualization of RenderNet's input consisting of G-Buffer (scene coordinates in camera space $X$, surface normals map $N$), material properties (albedo $A$, roughness $R$, specularity $S$), and lighting (light direction map $L_{dir}$, and light distance map $L_{dist}$).}
    
	\label{fig:geoemtric_scene_representation}  
\vspace{-5mm}
\end{figure}

As a first step, we define a geometric room representation outlining our synthetic environment. We place 3D objects inside an empty room ensuring no collisions. Next, we assign random materials to both objects and room walls and position a point light source to illuminate the scene (see Fig~\ref{fig:geoemtric_scene_representation}). Resulting output buffers, consisting of G-Buffer (scene coordinates in camera space $X$, surface normals map $N$), material properties (albedo $A$, roughness $R$, specularity $S$), and lighting (light direction map $L_{dir}$, and light distance map $L_{dist}$), are used by our neural ray tracer approximator to generate high fidelity renderings in real time ($\sim$2.5ms per image), as opposed to $\sim$4s per image with a conventional ray tracer.

% Given object masks and poses, and depth maps from a generated geometric representation, we obtain NOCS maps. To do that we first transform a depth map to a point cloud using camera parameters and transform each object instance using the inverse pose. The transformed normalized points are then stored as an image by utilizing initial pixel indices.

\subsection{Neural Ray Tracer Approximator}
\label{subsec:neural_ray_tracer}

Our neural ray tracer RenderNet $f_{R}$ is an encoder-decoder CNN taking G-buffer, material properties, and lighting as input, and generating a final high-fidelity rendering (see Fig.~\ref{fig:pndr}). This is akin to \emph{deferred rendering}, a common practice in computer graphics~\cite{deering1988triangle}. Instead of outputting a final rendering directly, we split the output into direct and indirect light outputs and colors which can be easily combined to form a final, shaded image. This allows not only for a much more explainable representation, but also for better control over the complexity of the rendering. As a result, our RenderNet $f_{R}$ is capable of generating photo-realistic images, generalizes well to novel material and light distributions, and even novel scenes, objects, and poses.  

% We will now explain the lighting and material properties that we employ, as well as our tone mapping.

\subsubsection{Light Modelling}

Lighting in ray tracers can often be decomposed into (1) direct lighting as coming from lamps, emitting surfaces, the background, or ambient occlusion after a single reflection or transmission from a surface; and (2) indirect lighting that comes from lamps, emitting surfaces or the background after more than one reflection or transmission. Simulating indirect lighting approximates realistic energy transfer much closer and produces better images, but comes at much higher computational cost.
To be computationally reasonable, we render all scenes with a single point light source.

\noindent\textbf{Light Sampler.} Our light sampler is a uniform random 3D coordinate generator. We limit the light pose space to the upper hemisphere and normalize the position to be at a distance of 1.5m from the scene center as defined in our training data. The resulting light source position in scene coordinates is then brought into the camera space given a fixed transform. Next, we parametrize the scene lighting by composing two light maps: $L_{dir}$ defines the direction to the light source from each visible coordinate and $L_{dist}$ defines the metric distance to the light source. Since RenderNet $f_{R}$ is fully differentiable, we can also use it to recover scene parameters in terms of lighting, particularly when combined with a correspondence-based object detector (see Sec.~\ref{subsec:6d_object_detector} with qualitative results in~\ref{subsec:image_editing}).  In this case we define the light sampling network $f_{L}$ as a SIREN-based MLP~\cite{sitzmann2020implicit} conditioned on the scene ID, which allows us to optimize for the light position given an input image.

% Since RenderNet $f_{R}$ is fully differentiable we can also produce augmentations using downstream loss guidance similar to ~\cite{zakharov2019deceptionnet}. In this case we define the light sampling network $f_{L}$ as a SIREN-based MLP~\cite{sitzmann2020implicit} conditioned on the scene ID and a noise value. We use a gradient reversal layer ~\cite{ganin2015unsupervised} as shown in Figure~\ref{fig:pndr} to revert the gradient for $f_{L}$ and adapt its weight to maximize a downstream task loss. 

\subsubsection{Material Modelling}

For both direct and indirect lighting our RenderNet $f_{R}$ outputs two separate images representing diffuse and glossy bidirectional scattering distribution functions (BSDF). The diffuse BSDF is used to add Lambertian~\cite{lambert1760photometria} and Oren-Nayar~\cite{oren1994generalization} diffuse reflection, whereas the glossy BSDF adds a GGX microfacet distribution~\cite{walter2007microfacet} that models metallic and mirror-like materials.

\noindent\textbf{Material Sampler.} Similarly to the light sampler, the material sampler is a uniform random value generator. It samples five values per object: RGB values for albedo $A$, roughness $R$ and specularity $S$ values. We query the material sampler for all objects in the scene including the background and, given ground truth instance masks, compose final 2D maps for each output property. The RGB albedo values are then multiplied by the GT decolorized albedo to form the final coloring map. Roughness and specular values are assigned to corresponding object masks to form full 2D maps.

Following a similar architecture to LightNet, we introduce an object material sampling network $f_{M}$, outputting material properties for each of the objects present in the dataset as well as the background environment. As shown in Fig.~\ref{fig:pndr}, it takes an object ID and scene ID values as input, and as before, produces the same object-specific material properties, i.e., albedo $A$, roughness $R$, and specularity $S$. Similarly to the uniform sampler, we query MaterialNet for all scene objects and compose final 2D maps for each output property.

\subsubsection{Image Compositing and Tone Mapping}
\label{subsec:tone_mapping}

Supplied with the G-buffer that provides us with scene coordinates in camera space $X$ and surface normals map $N$, we can form the final input to the RenderNet $f_{R}$ by concatenating all intermediate results and passing them through the encoder-decoder structure: 
\begin{equation}
    f_{R}(X, N, A, S, R, L_{dir}, L_{dist}) = [D_{dir}, D_{ind}, G_{dir}, G_{ind}].
\end{equation}
Here, $A, S, R, L_{dir}, L_{dist}$ are the outputs of the material and light submodules, as previously explained, whereas $D_{dir}, D_{ind}$ and $G_{dir}, G_{ind}$ are the diffuse and glossy BRDF outputs for direct and indirect lighting, respectively. During training we supervise the 4 outputs of RenderNet using corresponding ground truth quantities through an L1 loss. As outlined in Figure~\ref{fig:pndr}, the final HDR image is a combination of the light and BSDF outputs. In particular:
\begin{equation}
    I_{HDR} = (D_{dir} + D_{ind}) * D_{col} + (G_{dir} + G_{ind}) * G_{col},
\end{equation}
where D$_{col}$ represents object albedo and G$_{col}$ represents the probability that light is reflected for each wavelength. All compositing computations are performed in linear color space, which corresponds closer to nature and results in a more physically accurate output, i.e., there is a linear relationship between color intensity and the number of incident photons. However, these values do not directly correspond to human color perception and display devices. 
%That is why images are stored in different color spaces.

To address the limited color spectrum and brightness of displays, we apply a non-linear tone mapping to fit the device gamut. Although there are many families of mappings to choose from, we picked the commonly-used operator proposed by Jim Hejl and Richard Burgess-Dawson~\cite{hoffman2010crafting}. At its core, it is a rational function that mimics the response curve of a Kodak film commonly used in cinematography:

\begin{equation}
    I_{final} = \frac{I_{s}*(6.2*I_{s}+.5)}{I_{s}*(6.2*I_{s}+1.7)+0.06},
\end{equation}
where $I_{s} = max(0, I_{HDR}-0.004)$.

\section{Downstream Tasks}
\label{sec:downstream_tasks}

% \subsection{Correspondence-Based 6D Object Detection}
% \label{subsec:6d_object_detector}
We combine PNDR with two downstream task methods. At each training step, PNDR is used to generate photo-realistic augmentations on the fly, which are fed to the downstream task network. 

\begin{figure}[b!]
	\centering
	\includegraphics[width=1\linewidth]{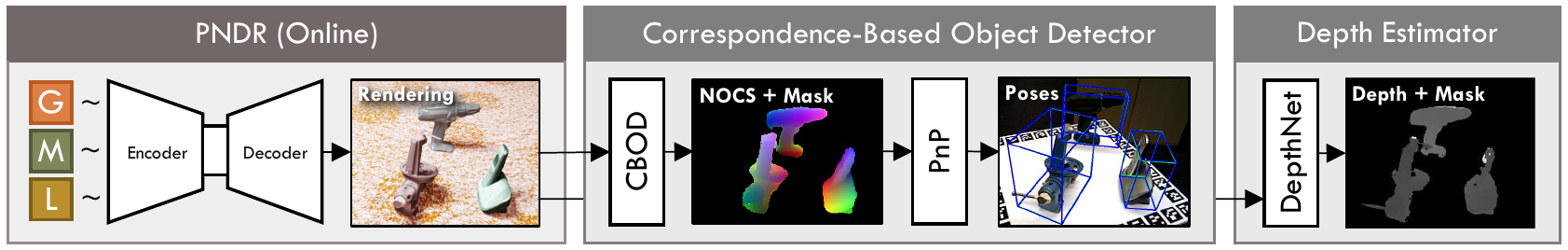}
	\caption{\textbf{Downstream Tasks Coupled with PNDR}. During training, both downstream tasks (detection and depth estimation) take PNDR renderings generated online, providing new realistic augmentations at each iteration.}
% 	Our general monocular correspondence-based detector takes an RGB images as an input and outputs non-uniform normalized object coordinates (NOCS) together with an ID mask. Estimated NOCS maps are used as correspondences to estimate final 6D poses for the detected objects. Similarly, our depth pipeline
% 		\ra{What would be really awesome here is (i) if the input, instead of an RGB image, shows the two approaches to generating an image: BlenderProc vs PNDR and (ii) the generated images are then fed into the two downstream task pipelines, Object detection and Monodepth.}
	\label{fig:cbod}  
\end{figure}

% We evaluate \textit{PNDR} on the task of 6D object detection.

\noindent\textbf{Correspondence-Based 6D Object Detection.} \label{subsec:6d_object_detector} 
Following~\cite{zakharov2019dpod,jafari2018ipose,park2019pix2pose,li2019cdpn,hodan2020epos}, our \textit{Correspondence-Based 6D Object Detector (CBOD)} operates on RGB images and outputs the probability of each pixel belonging to a certain local object coordinate. Estimated 2D-3D correspondences are then fed into a P\textit{n}P+RANSAC solver together with camera parameters to estimate final poses. We use non-uniform \textit{Normalized Object Coordinates Space (NOCS)}~\cite{wang2019normalized,zakharov2021single} maps that maximize the volume inside a unit cube, and we train using a cross-entropy loss. To disambiguate objects, our detector also outputs an instance segmentation mask. We then define object regions relying on instance mask probabilities and use respective correspondences to compute final poses. In addition to achieving competitive results, this simple yet effective architecture allows us to analyze the benefit of \textit{PNDR} not just for a single task of 6D pose estimation, but also for instance mask estimation and geometric correspondence accuracy, which are crucial components of general scene understanding. The structure of our detector is shown in Fig.~\ref{fig:cbod}, while the network architecture details are provided in the supplementary material.

\noindent\textbf{Monocular Depth Estimation.} 
We aim to learn a function $f_D:I \to D$ that recovers the depth $\hat{D}=f_D\left(I(p)\right)$ for every pixel $p\in I$. We operate in the supervised setting where we have access to the ground truth depth map, and we train the monocular depth network using the SILog loss~\cite{eigen2014depth,lee2019big} defined between the predicted and the ground truth depth maps. We evaluate the effect of PNDR using two network architectures: \textit{monodepth2}~\cite{godard2019digging} and \textit{packnet-sfm}~\cite{guizilini20203d}. 

\section{Experiments}
\label{sec:experiments}
We designed a number of experiments aimed at exploring how PNDR-generated data compares to real data as well as expensive, ray-tracer based simulation data in terms of downstream task performance.
% , time required to generate the synthetic data and train the model and storage space needed to store the generated data.

% \subsection{Datasets}
% \label{subsec:datasets}
% % \ra{Do we need this subsection?} \sz{Hmm, we could alternatively introduce them in the benchmark section}

% \noindent\textit{HomebrewedDB (HB)}~\cite{kaskman2019homebreweddb} is a recent dataset consisting of $13$ scenes of various complexity and featuring $33$ object models. We use publicly available labeled validation scenes for experiments, where each scene consists of $340$ frames that contain RGB images as well as associated object CAD models and exact 6D poses. 

% \paragraph{LineMOD (LM) \cite{hinterstoisser2012model}} is arguably one of the most popular benchmarks for evaluation of object detection and pose estimation methods. The original dataset consists of $13$ sequences, each containing ground truth poses for a single object of interest in a cluttered environment. CAD models for all the objects are provided as well.

\subsection{Evaluation Metrics}
\label{subsec:evaluation_metrics}

\noindent{\textbf{6D Object Detection.}} Following related work~\cite{zakharov2019dpod,kehl2017ssd}, we use ADD~\cite{hinterstoisser2012model} as the metric to evaluate object detection. ADD is defined as the average Euclidean distance between the model vertices transformed with ground truth and predicted poses:
\begin{equation}
\label{add_standard}
m = \avgunder_{\mathbf{x} \in \mathcal{M}} \norm{(\mathbf{R}\mathbf{x} + \mathbf{t}) - (\mathbf{\hat{R}}\mathbf{x} + \mathbf{\hat{t}})}_2,
\end{equation}
where  $\mathcal{M}$ is a set of vertices of a 3D model, ($\mathbf{R}, \mathbf{t}$) and ($\mathbf{\hat{R}},\mathbf{\hat{t}}$) are ground truth and predicted rotation and translation, respectively. Most commonly, a predicted pose is considered to be correct if ADD calculated with this pose is less than 10\% of a model diameter. However, this is a very strict metric especially for objects with a small diameter since it can completely disregard good pose estimates and estimates that could be refined. To be able to better analyze pose quality, we instead compute ADD under multiple thresholds (from 5 to 50 with a step of 5) and then estimate the area under the curve (AUC). 

\noindent\textbf{Instance Segmentation.} To evaluate the quality of the instance segmentation we use a standard Intersection over Union (IoU) metric, which quantifies the percent overlap between the target mask and our prediction output. 

\noindent\textbf{Object Correspondences.} To evaluate the quality of estimated object correspondences, we compare per-point metric distances in object's coordinate space between the GT partial shape and the predicted one. To do that we first use GT masks to recover partial object shapes and compute their absolute scale given provided model information. Then, we measure one-to-one distances between the GT shape and predicted shape in millimeters.

\noindent\textbf{Depth Estimation.} We evaluate the performance of our depth networks using the standard metrics found in the literature: \textit{AbsRel},  
\textit{RMSE} and \textit{$\delta_1$}, which are defined in detail in the supplementary.

\noindent\textbf{Perceptual Quality.} 
To evaluate generated RenderNet images, we use the standard image quality metrics
PSNR and SSIM~\cite{wang2004image} for all evaluations. Moreover, we include LPIPS~\cite{zhang2018unreasonable}, more accurately reflecting human
perception.

\section{Results}
\label{sec:results}

\begin{figure}[!b]
\minipage{0.44\textwidth}
	\includegraphics[width=1\linewidth]{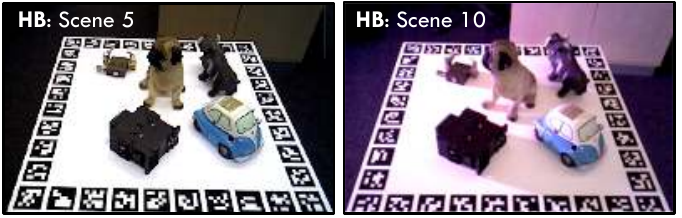}
	\caption{\textbf{HB Dynamic Lighting Benchmark:} two scenes containing same the objects under significantly different lighting conditions.}
	\label{fig:lighting}  
\endminipage\hfill
\minipage{0.54\textwidth}
	\includegraphics[width=1\linewidth]{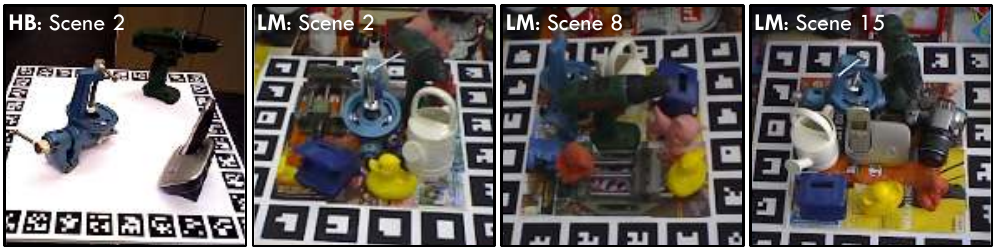}
	\caption{\textbf{HB-LM Cross-Domain Adaptation Benchmark:} four scenes containing the same objects in different environments and recorded with different cameras.}
	\label{fig:cross}  
\endminipage
\end{figure}

\subsection{HB Dynamic Lighting Benchmark}
\label{subsec:dynamic_lighting_benchmark}

\setlength{\tabcolsep}{4pt}
\begin{table*}[t]
  \centering
    \resizebox{1\linewidth}{!}{
    \begin{tabular}{c|c|ccccc|c|ccccc|c}
    \toprule
    \multirow{2}[4]{*}{\textbf{Train}} & \multirow{2}[4]{*}{\textbf{Method}} & \multicolumn{6}{c|}{\textbf{HB Scene 5}}         & \multicolumn{6}{c}{\textbf{HB Scene 10 (Lighting)}} \\
\cmidrule{3-14}          &       & \textbf{Car} & \textbf{P12} & \textbf{P15} & \textbf{Pumba} & \textbf{Dog} & \textbf{Mean} & \textbf{Car} & \textbf{P12} & \textbf{P15} & \textbf{Pumba} & \textbf{Dog} & \textbf{Mean} \\
    \midrule
    \multicolumn{1}{c|}{Real} &       & 91.76 & 94.12 & 79.34 & 94.41 & 95.00 & 90.93 & 26.69 & 35.88 & 8.75  & 25.22 & 14.63 & 22.24 \\
    \midrule
    \multicolumn{1}{c|}{\multirow{2}[2]{*}{\makecell{Real \\+ CAD$\dagger$}}} & Pix2Pix~\cite{isola2017image} & 81.84 & 76.47 & 38.53 & 76.76 & 94.12 & 73.54 & 37.13 & 35.59 & 13.75 & 24.78 & 33.82 & 29.01 \\
          & Pix2PixHD~\cite{wang2018high} & 90.96 & 92.21 & 73.09 & 91.91 & 95.00 & 89.59 & 37.21 & 40.96 & 21.25 & 29.26 & 15.59 & 32.33 \\
    \midrule
    \multicolumn{1}{c|}{\multirow{2}[2]{*}{\makecell{Real \\+ CAD$\ddagger$}}} & CycleGAN~\cite{zhu2017unpaired} & 49.41 & 31.10 & 24.12 & 40.88 & 73.24 & 43.75 & 27.72 & 2.28  & 6.91  & 7.06  & 11.47 & 11.09 \\
          & CUT~\cite{park2020contrastive}   & 56.10 & 28.97 & 29.34 & 41.47 & 85.29 & 66.49 & 27.35 & 4.63  & 7.35  & 8.90  & 12.13 & 25.36 \\
    \midrule
    \multicolumn{1}{c|}{\multirow{2}[2]{*}{CAD}}
    % DR &       &       &       &       &       &       &       &       &       &       &       &  \\
    % & Phong &       &       &       &       &       &       &       &       &       &       &       &  \\
    & RayTraced - 1088 & 85.59 & 86.76 & 61.18 & 89.71 & 94.85 & 83.62 & 47.28 & 36.84 & 9.12 & 36.25 & 23.38 & 30.57 \\
    & RayTraced - 2176 & 86.99 & 90.00 & 63.01 & 91.47 & 95.00 & 85.29 & 50.88 & 38.82 & 10.00  & 35.29 & 30.96 & 33.19 \\
    & RayTraced - 4352  & 89.71 & 88.97 & 66.91 & 92.35 & 95.00 & 86.59 & 52.43 & 38.75 & 10.29  & 41.47 & 43.82 & 37.35 \\
    & Ours - 1088 & 89.93 & 91.62 & 71.99 & 92.35 & 95.00 & 88.18 & 58.01 & 42.50 & 10.59 & 46.18 & 44.93 & 40.44 \\
        %   & Ours$_{G}$ &       &       &       &       &       &       &       &       &       &       &       &  \\
    \bottomrule
    \end{tabular}%
    }
    % \vspace{2mm}
    \caption{\textbf{HB Dynamic Lighting Benchmark}: All methods are trained on the training set of $HB_5$ and evaluated on the $HB_5$ test set and on $HB_{10}$. $\dagger$ indicates that~\cite{isola2017image,wang2018high} are trained with synthetic and real image pairs, while $\ddagger$ indicates unpaired synthetic and real images for~\cite{zhu2017unpaired,park2020contrastive}. Training on photo-realistic synthetic data is competitive with real data training and generalizes better to new domains. By training on \textit{PNDR} images we further close the gap to training on real data in $HB_5$ and increase generalization performance to the novel lighting setting of $HB_{10}$.}
  \label{tab:lighting}
\end{table*}%

In this first experiment we aim to isolate the effects of training and testing under significantly different illumination while keeping the scene contents constant. We use scenes 5 and 10 of the \textit{HomeBrewedDB} (HB) dataset~\cite{kaskman2019homebreweddb} ($HB_5$ and $HB_{10}$ for short), as shown in Fig.~\ref{fig:lighting}. Both scenes contain the same objects in the same environment and consist of 340 images with associated depth maps and object annotations (CAD models and poses). $HB_5$ and $HB_{10}$ are captured with drastically different lighting conditions, allowing us to isolate the effect simulated data has on overcoming this perceptual domain gap. We split $HB_5$ into a training and in-domain testing subsets consisting of 272 and 68 frames respectively; $HB_{10}$ is used entirely for testing. 

\begin{figure*}[t]
	\centering
	\includegraphics[width=.98\linewidth]{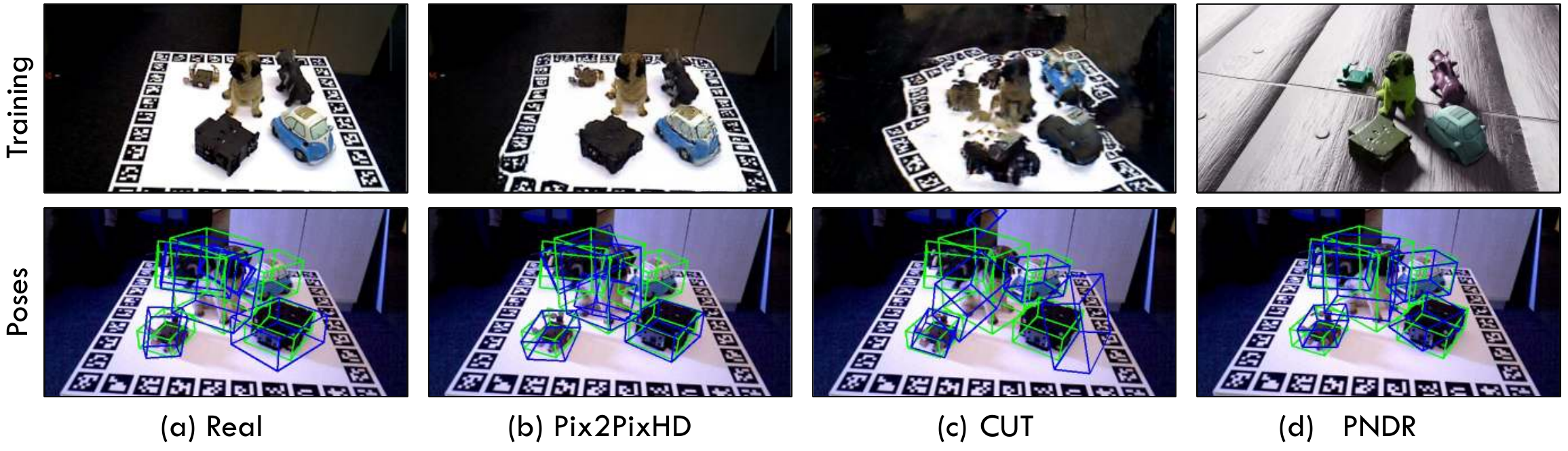}
	\caption{\textbf{6D Object detection qualitative results.} We compare object detection results when trained on \textit{PNDR} renderings with our image-to-image translation GAN baselines and with the baseline trained on real data. All methods are evaluated on an $HB_5$ test image.}
	\label{fig:baselines}  
\end{figure*}

% \noindent\textbf{6D Object Detection using CBOD.} 
We present the benchmark results in Table~\ref{tab:lighting} (qualitative results are shown in Fig.~\ref{fig:baselines}). Our first baseline consists of training CBOD directly on the $HB_5$ real images, and we record good in-domain performance (90.93) and poor transfer to different light configurations (22.24). Our second baseline uses entirely synthetic photo-realistic images of increasing sizes. Using the object CAD models and associated poses corresponding to the different training frames (i.e. we have a total of 272 different object configurations), we generate Domain Randomized synthetic photo-realistic images with \textit{BlenderProc}~\cite{denninger2019blenderproc}. Specifically, for each training  configuration  we vary object materials and light positions. For backgrounds we randomly select from 5 different asset classes (\textit{Bricks}, \textit{Wood}, \textit{Carpet}, \textit{Tile}, \textit{Marble}~\footnote{\url{https://ambientcg.com}}) and also randomize their materials. For each training configuration we generate an increasing number of augmentations using this technique, leading to larger synthetic datasets with very high perceptual quality at the expense of rendering time and storage space. We train CBOD on the synthetic images, and, as expected, downstream task performance improves as more high-quality synthetic data is available (i.e., with 4352 synthetic images we achieve 37.35 generalization performance). We compare this with the proposed PNDR method as follows: using the high-quality synthetic data along with the corresponding G-buffer information, we train PNDR, and use it in the training pipeline of CBOD to generate new, high-quality augmentations on the fly, saving rendering time and storage space. As shown in our experiments, 1088 synthetic images are enough to train PNDR, and we almost match the performance of training on real data and increase generalization to scenes with significant light variation by $82\%$ ($40.44$ vs $22.24$). We note that as CBOD is trained over 400 epochs, it would require $\sim$30h and 600GB storage space to generate as many images with the raytracer as were generated by PNDR.

Additionally we train state-of-the-art image-translation methods on the same images we train \textit{PNDR} on, and we use the corresponding real images as translation targets. Specifically, we compare against (i) pix2pix~\cite{isola2017image} and pix2pixHD~\cite{wang2018high} using BlenderProc and real image pairs; and (ii) cycleGAN~\cite{zhu2017unpaired} and CUT~\cite{park2020contrastive} using unpaired BlenderProc and real images. Although our method does not require any real images, adversarial image translation methods are representative of the state-of-the-art in domain adaptation and serve as good baselines. The training details of all the baselines are provided in the supplementary.  The paired GAN baselines~\cite{isola2017image,wang2018high} increase generalization performance, although we note that having access to synthetic and image pairs is an unrealistic scenario in practice and this serves as an upper-bound, at least for in-domain evaluation. The more realistic case of unpaired translation~\cite{zhu2017unpaired,park2020contrastive} performs much worse, as expected. This is easily explained by the rather large domain shift induced from the different scene setups and cameras.

\subsection{HB-LM Cross-Domain Adaptation Benchmark}
\label{subsec:HB_LM_cross_domain}

\setlength{\tabcolsep}{8pt}
\begin{table*}[t]
  \centering
  
    \resizebox{1\linewidth}{!}{
    \begin{tabular}{c|c|ccc|c|ccc|c}
    \toprule
    \multirow{2}[4]{*}{\textbf{Train}} & \multirow{2}[4]{*}{\textbf{Method}} & \multicolumn{4}{c|}{\textbf{HB Scene 2}} & \multicolumn{4}{c}{\textbf{LM Scenes 2, 8, 15}} \\
\cmidrule{3-10}          &       & \textbf{Bvise} & \textbf{Drill} & \textbf{Phone} & \textbf{Mean} & \textbf{Bvise} & \textbf{Drill} & \textbf{Phone} & \textbf{Mean} \\
    \midrule
    \multicolumn{1}{c|}{Real} &       & 94.71 & 95.00 & 94.41 & 94.71 & 4.43  & 0.30  & 0.35  & 1.69 \\
    \midrule
    \multicolumn{1}{c|}{\multirow{2}[2]{*}{\makecell{Real + CAD$\dagger$}}} & Pix2Pix~\cite{isola2017image} & 75.66 & 77.94 & 67.65 & 73.75 & 4.97  & 1.38  & 0.28  & 2.21 \\
          & Pix2PixHD~\cite{wang2018high} & 92.57 & 94.41 & 91.76 & 92.92 & 4.65  & 1.62  & 0.47  & 2.25 \\
    \midrule
    \multicolumn{1}{c|}{\multirow{2}[2]{*}{\makecell{Real + CAD$\ddagger$}}} & CycleGAN~\cite{zhu2017unpaired} & 35.37 & 26.76 & 45.22 & 35.78 & 5.68  & 5.40  & 3.23  & 4.77 \\
          & CUT~\cite{park2020contrastive}  & 78.53 & 61.25 & 65.44 & 68.41 & 21.70 & 10.05 & 6.33  & 12.69 \\
    \midrule
    \multicolumn{1}{c|}{\multirow{2}[2]{*}{CAD}} 
    % & DR &       &       &       &       &       &       &       &  \\
    % & Phong &       &       &       &       &       &       &       &  \\
    & RayTraced - 1088  & 84.49 & 72.43 & 80.81 & 79.24 & 33.67 & 9.55  & 15.90 & 19.71 \\
    & Ours - 1088 & 85.88 & 81.54 & 83.09 & 83.50 & 35.50 & 28.02 & 18.48 & 27.33 \\
        %   & Ours$_{G}$ &       &       &       &       &       &       &       &  \\
    \bottomrule
    \end{tabular}%
    }
    % \vspace{1mm}
    \caption{\textbf{HB-LM Cross-Domain Adaptation Benchmark}: All methods are trained on the training set of $HB_2$ and evaluated on $HB_2$ test set as well as on the $LM_2$, $LM_8$ and $LM_{15}$ scenes. Training on real data generalizes poorly to novel object categories. Training on photo-realistic synthetic data is competitive with real data training when evaluated on the same object categories, and greatly increases generalization performance for novel object categories. $\dagger$ indicates that~\cite{isola2017image,wang2018high} are trained with paired synthetic and real image pairs, while $\ddagger$ indicates unpaired images for~\cite{zhu2017unpaired,park2020contrastive}.}
  \label{tab:cross}
%   \vspace{-3mm}
\end{table*}%

%  \begin{figure*}[t]
% 	\centering
% 	\includegraphics[width=1\linewidth]{figures/rendernet.pdf}
% 	\caption{\textbf{PNDR generalization and sampling}. Our method is trained on $HB_5$ training set. Left: the top row shows \textit{PNDR} performance on $HB_5$ test images, while the bottom row shows \textit{PNDR} performance on $HB_2$ which contains unseen objects. Right: \textit{PNDR} generates photo-realistic renderings by sampling materials and light positions, both for objects seen during training as well as for new object categories.}
% 	\label{fig:pndr_qualitative}  
% \end{figure*}

% \noindent\textbf{HB-LM Cross-Domain Adaptation Benchmark} 
Our HB-LM cross-domain benchmark (see Fig.~\ref{fig:cross}) is represented by $HB_2$ covering three objects of the LineMOD (LM)~\cite{hinterstoisser2012model}  dataset (\textit{benchvise}, \textit{driller}, and \textit{phone}). Additionally, we use scenes 2, 8 and 15 from the LM dataset for testing: these scenes contain the same objects as $HB_2$ but with significantly different poses and in a different setting. This setting allows us to evaluate the generalization performance of \textit{PNDR} to new scenes and new object poses. As before, we partition the $HB_2$ into a training and test split consisting of 272 and 68 images, respectively, and we use \textit{BlenderProc}~\cite{denninger2019blenderproc} to generate the same synthetic photo-realistic renderings and G-buffer information. In addition to the $HB_2$ data, we also generate 1000 photo-realistic images using \textit{BlenderProc} while randomizing both the camera and the poses of the 3 objects from $HB_2$. As we show in the experiments, using this extra simulated data allows us to generalize much better to the $LM$ scenes where the object pose distribution is significantly different. We train \textit{PNDR} as before and use its output to train \textit{CBOD}, which is evaluated both in domain, i.e., on the test split of $HB_2$ as well as out of domain on $LM_2$, $LM_8$ and $LM_{15}$.

We analyze how well we generalize to completely different scenes with different lighting conditions, environment, camera setup and object poses; our results are summarized in Table~\ref{tab:cross}. As before, we report the best in-domain results when training on real data, with a slight performance drop when training directly on the photo-realistic synthetic BlenderProc data. We note that by using \textit{PNDR} we significantly increase performance. As before, the paired translation GAN baselines compare quite well, and we record a similar performance drop when doing unpaired image translation. Both the GAN and the baseline trained on real data generalize poorly to the LM scenes, reflecting the challenging nature of this benchmark. Interestingly, the unpaired image translation baselines generalize better in this setting - we provide qualitative examples in the supplementary. By training on the synthetic data which contains additional renderings with randomized object poses we significantly improve performance. As before, using \textit{PNDR} as part of the training pipeline further improves performance, achieving $27.33$ on the LM scenes.

\subsection{HB Generalization Benchmark}
\label{subsec:renderer_generalization}

\begin{figure}[b]
	\centering
	\includegraphics[width=1\linewidth]{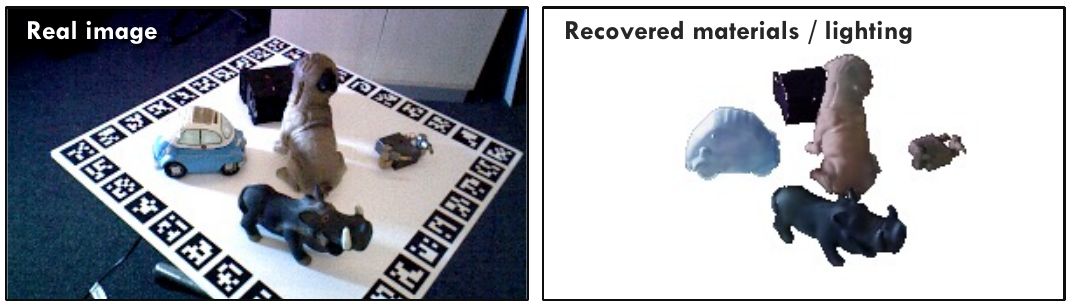}
	\caption{\textbf{Recovering scene properties via our RenderNet}. Given the recovered G-Buffer we optimize over \textit{MaterialNet} and \textit{LightNet} to find the best fit explaining the input image.}
	\label{fig:image_editing}  
\end{figure}

% \noindent\textbf{HB Generalization Benchmark.} 
Here we aim to evaluate how well \textit{PNDR} can generate novel photo-realistic images in and out of domain. We train \textit{PNDR} on $HB_2$ and evaluate on the test set of $HB_2$ as well as $HB_5$. Note that these two scenes contain different objects, allowing us to investigate if our learned ray-tracing network generalizes to novel scene geometries. For completeness, we also perform the same experiment by training on $HB_5$ and evaluating on $HB_2$.

We quantify the generalization capabilities of \textit{PNDR} when applied to novel scenes, object arrangements, material properties and lighting. We train \textit{PNDR} on the training split of $HB_2$ and evaluate how it performs when applied to: (1) the training split, (2) the test split, (3) the test split of a different scene (i.e., $HB_5$). For completeness, we also perform the reverse experiment (i.e., train on $HB_5$ and evaluate on $HB_2$). The results in Table~\ref{tab:rendering} suggest a high level of visual quality when assessed with PSNR, SSIM, and LPIPS (please refer to the supplementary for details on these metrics). \textit{PNDR} not only shows strong results on the test set containing the same objects under different poses and with different material properties, but also generalizes well to a scene with completely different objects. 
% We present qualitative results in Fig.~\ref{fig:pndr_qualitative}.

\begin{table}[t]
\parbox[t]{.5\linewidth}{
  \centering
    \resizebox{1\linewidth}{!}{
    \begin{tabular}{c|c|ccc}
    \toprule
    \textbf{Train} & \textbf{Test} & \textbf{PSNR$\uparrow$} & \textbf{SSIM$\uparrow$} & \textbf{LPIPS$\downarrow$} \\
    \midrule
    \multirow{3}[1]{*}{$HB_2$ - Train} & $HB_2$ - Train & 30.36 & 0.96  & 0.03 \\
          & $HB_2$ - Test  & 26.14 & 0.94  & 0.05 \\
          & $HB_5$ - Test  & 24.14 & 0.92  & 0.06 \\
    \midrule
    \multirow{3}[1]{*}{$HB_5$ - Train} & $HB_5$  - Train & 30.39 & 0.96  & 0.03 \\
          & $HB_5$ - Test  & 26.40 & 0.94  & 0.06 \\
          & $HB_2$ - Test  & 23.72 & 0.92  & 0.07 \\
    \bottomrule
    \end{tabular}%
    }
    % \vspace{1mm}
    \caption{\textbf{PNDR Generalization:} We achieve strong performance not only when applied to a test image of the same scene containing the same objects with different material properties and different poses, but also when applied on a completely different scene.}
  \label{tab:rendering}
}
\hfill
\parbox[t]{.46\linewidth}{
  \centering
    \resizebox{1\linewidth}{!}{
    \begin{tabular}{r|r|ccc}
    \toprule
          & \multicolumn{1}{c|}{\textbf{Modes}} & \textbf{ADD$_{\textbf{AUC}}$} $\uparrow$ & \textbf{IoU}  $\uparrow$ & \textbf{Corr (mm)  $\downarrow$} \\
    \midrule
    \multirow{2}[2]{*}{Material} & \multicolumn{1}{c|}{$A$} & 85.43 & 83.29 & 38.56 \\
          & \multicolumn{1}{c|}{$+S+R$} & 3\%   & 3\%   & 11\% \\
    \midrule
    \multirow{2}[2]{*}{Light} & \multicolumn{1}{c|}{Fixed} & 73.04 & 72.24 & 76.94 \\
          & \multicolumn{1}{c|}{Dynamic} & 21\%  & 19\%  & 55\% \\
    \midrule
    \multirow{2}[2]{*}{Rendering} & \multicolumn{1}{c|}{$D_{dir}$, $G_{dir}$} & 87.09 & 85.55 & 38.42 \\
          & \multicolumn{1}{c|}{+ $D_{ind}$, $G_{ind}$} & 1\%   & 0\%   & 11\% \\
    \midrule
    Full  &       & 88.18 & 85.97 & 34.32 \\
    \bottomrule
    \end{tabular}%
    }
    % \vspace{1mm}
    \caption{\textbf{Ablation:} We analyze the effect of different augmentations on downstream task performance. All methods are trained on the $HB_5$ train set and evaluated on the $HB_5$ test set.}
  \label{tab:ablation}%
}
% \vspace{-6mm}
\end{table}

\subsection{Ablation Study}
\label{subsec:ablation}
We analyze how different physically-based augmentations affect the downstream task performance, and consider: material randomization, light randomization, and rendering complexity. \textit{PNDR} is conditioned on material properties of the objects, i.e., albedo $A$, specularity $S$, and roughness $R$, with $A$ being the most important property. In Table~\ref{tab:ablation} we see that training with just albedo randomization results in very good performance already. Additionally simulating $S$ and $R$ brings a relative gain of 2\% with respect to ADD$_{\text{AUC}}$, 3\% to mIoU, and 11\% to correspondence quality. Furthermore, we note that lighting is by far the most important randomization parameter. Going from fixed to dynamic lighting significantly improves the results: 19\% ADD$_{\text{AUC}}$ gain, 19\% mIoU gain, and 55\% correspondence quality gain. Finally, we note that simulating computationally expensive indirect lighting only helps improve correspondence quality but is negligible for the other metrics. Since we use an outlier-robust P\textit{n}P+RANSAC solver, small deviations in correspondence quality do not significantly affect the pose quality as evaluated by ADD$_{\text{AUC}}$.

\subsection{Monocular Depth Estimation} 
We quantify the impact of PNDR when applied for the task of monocular depth estimation. We use the same data as for the Dynamic Lighting Benchmark (see~\ref{subsec:dynamic_lighting_benchmark}). For both \textit{monodepth2} and \textit{packnet-sfm} we compare performance when training directly on the 1088 raytraced images with performance when PNDR is integrated in the training pipeline and generates novel augmentations on the fly. As before, we note that when training with PNDR we achieve better in-domain and better generalization performance (see Table~\ref{table:monodepth}). 
% \ra{Should this section go somewhere else, and we describe object detection first for all the benchmarks?} \sz{Hmm, yes, since it's a secondary task, we should probably put it after detection?}

\setlength{\tabcolsep}{10pt}
\begin{table}[t]
  \centering
%   \caption{Add caption}
    \resizebox{1\linewidth}{!}{
    \begin{tabular}{c|c|ccc|ccc}
    \toprule
    \multirow{2}[4]{*}{\textbf{Method}} & \multirow{2}[4]{*}{\textbf{Training}} & \multicolumn{3}{c|}{\textbf{HB5}} & \multicolumn{3}{c}{\textbf{HB10}} \\
\cmidrule{3-8}          &       & \textbf{AbsRel}$\downarrow$ & \textbf{RMSE}$\downarrow$ & \textbf{a1}$\uparrow$ & \textbf{AbsRel}$\downarrow$ & \textbf{RMSE}$\downarrow$ & \textbf{a1}$\uparrow$ \\
    \midrule
    \multirow{2}[2]{*}{Monodepth2} & Raytraced & 0.082 & 0.09  & 0.951 & 0.162 & 0.148 & 0.805 \\
          & PNDR  & \textbf{0.075} & \textbf{0.083} & \textbf{0.966} & \textbf{0.154} & \textbf{0.14} & \textbf{0.83} \\
    \midrule
    \multirow{2}[2]{*}{PackNet-SfM} & Raytraced & 0.11  & 0.111 & 0.884 & 0.141 & 0.136 & 0.833 \\
          & PNDR  & \textbf{0.082} & \textbf{0.087} & \textbf{0.977} & \textbf{0.135} & \textbf{0.131} & \textbf{0.852} \\
    \bottomrule
    \end{tabular}%
    }
    % \vspace{1mm}
    \caption{
\textbf{PNDR vs Raytraced - monocular depth results.}
% \vspace*{-3mm}
}
  \label{table:monodepth}%
\end{table}% 

\subsection{Object Material and Light Recovery}
\label{subsec:image_editing}
The fact that \textit{RenderNet} is fully differentiable allows us to optimize over scene parameters. In particular, given an initial scene prediction provided by \textit{CBOD} we can recover material properties of the objects and scene light (see Fig.~\ref{fig:image_editing}). First, we construct a G-buffer by estimating a depth map using estimated poses and object models, which is in turn used to estimate scene coordinates $X$ and surface normals $N$. LightNet $f_L$ and MaterialNet $f_M$ conditioned on the scene and object IDs output the remaining maps $A$, $R$, $S$, $L_{dir}$, and $L_{dist}$ required for the RenderNet. Finally we generate a rendering that is then compared to the GT RGB image to find the best fit. Estimated scene parameters can be used to learn a distribution of material and light configurations across the entire dataset. This information might not only be useful for analysis, but also for domain-specific data generation using our RenderNet, especially where the same object instances exist in multiple material variations.

\section{Conclusion}
We have presented a novel approach towards sim-to-real adaptation by means of a neural ray tracer approximator with randomizable material and light modules that we named \textit{PNDR}. We have demonstrated that applying our photo-realistic randomized output to the problem of zero-shot 6D object detection significantly outperforms other established DA approaches, and even comes close to training on real data. We have identified lighting as the most crucial component, but it remains an open question what kind of additional randomization could further benefit the domain transfer. One possible future research avenue would be randomized sampling of low-level camera sensor artifacts, or the coupling of randomization and downstream task optimization in a common framework.

\clearpage
\bibliographystyle{splncs04}
\bibliography{egbib}

\clearpage
\appendix
\section{Supplementary Material}
\subsection{RenderNet}
\label{sec:supp_PNDR_networks}
Our RenderNet is a UNet-based encoder-decoder CNN. It takes a 15D input (see Fig. 1 of the main submission) consisting of concatenated scene coordinates in camera space $X$ (3D), surface normals map $N$ (3D), albedo $A$ (3D), roughness $R$ (1D), specularity $S$ (1D), light direction map $L_{dir}$ (3D), and light distance map $L_{dist}$ (1D). Its inference results in four 3D maps: $D_{dir}, D_{ind}, G_{dir}$, and $G_{ind}$ being the diffuse and glossy BSDF outputs for direct and indirect lighting, respectively. Predicted outputs are then used to form a final rendering. 
% For examples of both RenderNet's inputs and outputs see Figure~\ref{}.

To train it, we used the Adam optimizer~\cite{kingma2014adam} with a learning rate of $1e^{-4}$. Each of the four RenderNet outputs is supervised with respective ground truth images using an L1 loss.

% \subsection{Training details}
% \label{subsec:supp_PNDR_training_details}

\subsection{CBOD}
\label{sec:supp_CBOD_networks}

Our CBOD detector consists of two modules: the correspondence module and the pose estimation module. This section provides their detailed description. Our detector largely follows~\cite{zakharov2019dpod,jafari2018ipose,park2019pix2pose,li2019cdpn,hodan2020epos}.

\subsubsection{Correspondence Module.} Our correspondence module is a ResNet12-based encoder-decoder CNN with four decoder heads to regress the ID mask and three channels of the dense 2D-3D correspondence map (U, V, W) from a 320$\times$240$\times$3 RGB image. However, we would like to note that any other backbone architecture could be used without any need to change the rest of the pipeline. The decoders upsample features up to their original size using a stack of bilinear interpolations followed by convolutional layers. 

\begin{figure}[b!]
	\centering
	\includegraphics[width=1\linewidth]{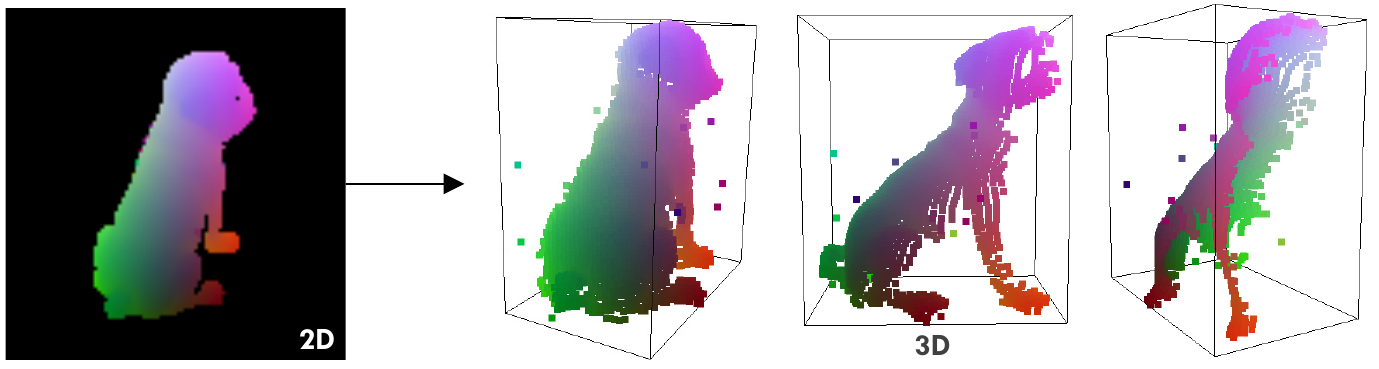}
	\caption{\textbf{2D-3D correspondences.} We recover partial geometry from a regressed 2D NOCS map and establish 2D-3D correspondences.}
	\label{fig:nocs}  
\end{figure}

Correspondence heads regress tensors of size  $H$$\times$$W$$\times$$C$, where $C$ is the discretization density of the correspondence map, which equals to 256 in our case. Each channel stores the probability values for the class corresponding to a specific channel number. Once regressed, we compose single channel tensors (U, V, W) storing the class ID with maximal probability using the argmax operation. Defining correspondence estimation as a classification problem allows us to significantly decrease the output solution space, which subsequently results in a better quality of 2D-3D matches and faster convergence. Resulting U, V, and W channels are used to form a 2D NOCS map encoding normalized object's coordinates in RGB. Visualizing each color component in 3D allows us to restore a partial object's geometry as can be seen in Fig.~\ref{fig:nocs} and establish 2D-3D correspondences needed for pose estimation.

Similarly, the ID mask head outputs a $H$$\times$$W$$\times$$O$ tensor with $O$ corresponding to the number of objects in the dataset plus one additional class for background. We apply an argmax operation to the output tensor and identify detected classes. Then, we use detected classes to retrieve class-specific channels, apply Otsu's method~\cite{otsu1979threshold} to retrieve detected masks, and finally identify connected mask regions.  This strategy proved to be more robust when compared to directly using argmax-based masks to define regions. Resulting mask regions are then used by the pose module.

The final loss function for \textit{RenderNet} is defined as the sum of four losses:
\begin{equation}
\mathcal{L} = \mathcal{L}_m + \mathcal{L}_u + \mathcal{L}_v + \mathcal{L}_w,
\end{equation}
where $\mathcal{L}_m$ is the mask loss, and $\mathcal{L}_u$, $\mathcal{L}_v$, and $\mathcal{L}_w$ are the losses responsible for the separate correspondence map channels U, V, and W. All losses are defined as multi-class cross-entropy functions.

% \begin{figure*}[t]
% 	\centering
% 	\includegraphics[width=1\linewidth]{figures/supplementary/input.pdf}
% 	\caption{\textbf{RenderNet's input.} Visualization of the full RenderNet's input consisting of G-Buffer (scene coordinates in camera space $X$, surface normals map $N$), material properties (albedo $A$, roughness $R$, specularity $S$), and lighting (light direction map $L_{dir}$, and light distance map $L_{dist}$).}
% 	\label{fig:input}  
% \end{figure*}

\begin{figure*}[b]
	\centering
	\includegraphics[width=1\linewidth]{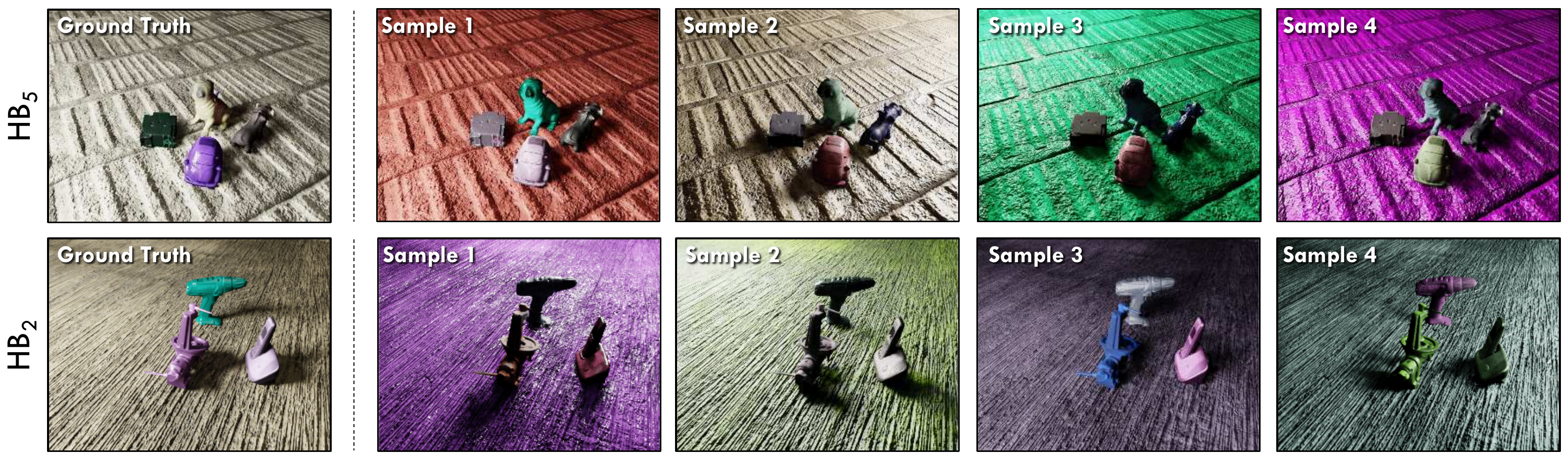}
	\caption{\textbf{Sampling novel light and materials.} \textit{PNDR} generates photo-realistic renderings by sampling materials and light positions.}
	\label{fig:sampling}  
\end{figure*}

\subsubsection{Pose Module.}
Given detected regions and estimated 2D-3D correspondences, we use a Perspective-\textit{n}-Point (P\textit{n}P)~\cite{quan1999linear} solver that estimates the camera pose given correspondences and intrinsic parameters of the camera.
Random sample consensus or RANSAC~\cite{fischler1981random} is used in conjunction with P\textit{n}P to make predictions more robust to outliers. We use a standard P\textit{n}P-RANSAC implementation Perspective-\textit{n}-Point (P\textit{n}P) provided in the OpenCV function \texttt{solvePnPRansac}. We set the number of RANSAC iterations to 300 and reprojection error threshold to $1$.

% \subsection{Training details}
% \label{subsec:supp_CBOD_training_details}
\subsubsection{Training Details.}
To train CBOD, we used the ADAM optimizer~\cite{kingma2014adam} with a learning rate of $5e^{-4}$ and weight decay of $4e^{-5}$. As opposed to \cite{Hinterstoisser2017,Manhardt2018,zakharov2019dpod}, we do not use pretrained models and do not freeze the first layers of the network to have a fair comparison between different types of data.

\subsection{Baselines}
\label{sec:supp_baselines}

\begin{figure}
\captionsetup[subfigure]{labelformat=empty,font=scriptsize}
    \centering
    \subfloat{
    \includegraphics[width=0.15\textwidth]{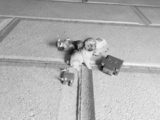}}
    \subfloat{
    \includegraphics[width=0.15\textwidth]{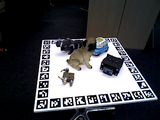}}
    \subfloat{
    \includegraphics[width=0.15\textwidth]{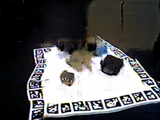}}
    \subfloat{
    \includegraphics[width=0.15\textwidth]{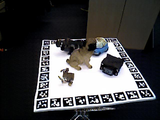}}
    \subfloat{
    \includegraphics[width=0.15\textwidth]{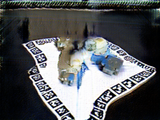}}
    \subfloat{
    \includegraphics[width=0.15\textwidth]{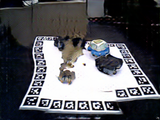}}
    \\ 
    \subfloat{
    \includegraphics[width=0.15\textwidth]{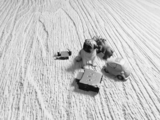}}
    \subfloat{
    \includegraphics[width=0.15\textwidth]{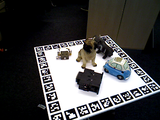}}
    \subfloat{
    \includegraphics[width=0.15\textwidth]{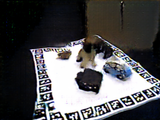}}
    \subfloat{
    \includegraphics[width=0.15\textwidth]{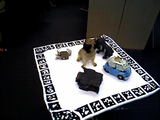}}
    \subfloat{
    \includegraphics[width=0.15\textwidth]{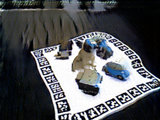}}
    \subfloat{
    \includegraphics[width=0.15\textwidth]{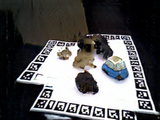}}
    \\ 
    \setcounter{subfigure}{0}
    \subfloat[][Input synthetic image]{
    \includegraphics[width=0.15\textwidth]{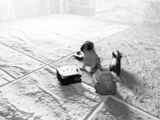}}
    \subfloat[Real image]{
    \includegraphics[width=0.15\textwidth]{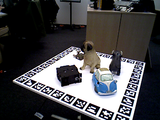}}
    \subfloat[pix2pix~\cite{isola2017image}]{
    \includegraphics[width=0.15\textwidth]{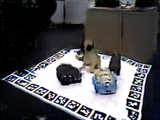}}
    \subfloat[][pix2pixHD~\cite{wang2018high}]{
    \includegraphics[width=0.15\textwidth]{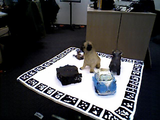}}
    \subfloat[][cycleGAN~\cite{zhu2017unpaired}]{
    \includegraphics[width=0.15\textwidth]{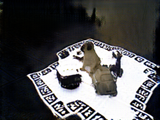}}
    \subfloat[CUT~\cite{park2020contrastive}]{
    \includegraphics[width=0.15\textwidth]{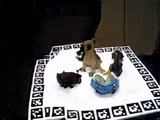}}
    \caption{\textbf{Qualitative results on HB5.} We show the results of the paired and upaired image translation baselines on the HB5 scene used in the HB Dynamic Lighting Benchmark. Top row: images from the $HB_5$ training set; last two rows: images from the $HB_5$ test set. The real image is used as pair target when training ~\cite{isola2017image,wang2018high} while~\cite{zhu2017unpaired,park2020contrastive} use an unordered set of real images as targets.}
    \label{fig:baselines_qualitative_HB_lighting}
\end{figure}

\begin{figure*}
\captionsetup[subfigure]{labelformat=empty,font=scriptsize}
    \centering
    \subfloat{
    \includegraphics[width=0.15\textwidth]{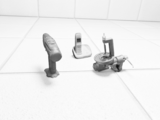}}
    \subfloat{
    \includegraphics[width=0.15\textwidth]{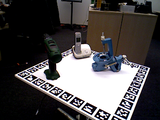}}
    \subfloat{
    \includegraphics[width=0.15\textwidth]{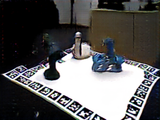}}
    \subfloat{
    \includegraphics[width=0.15\textwidth]{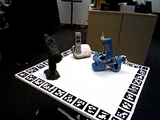}}
    \subfloat{
    \includegraphics[width=0.15\textwidth]{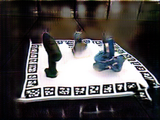}}
    \subfloat{
    \includegraphics[width=0.15\textwidth]{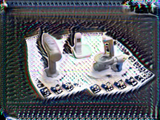}}
    \\ 
    \subfloat{
    \includegraphics[width=0.15\textwidth]{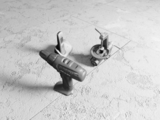}}
    \subfloat{
    \includegraphics[width=0.15\textwidth]{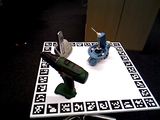}}
    \subfloat{
    \includegraphics[width=0.15\textwidth]{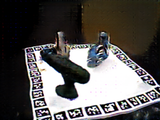}}
    \subfloat{
    \includegraphics[width=0.15\textwidth]{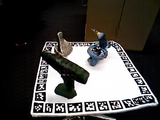}}
    \subfloat{
    \includegraphics[width=0.15\textwidth]{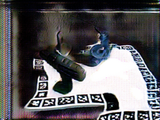}}
    \subfloat{
    \includegraphics[width=0.15\textwidth]{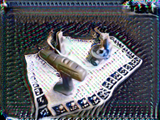}}
    \\ 
    \setcounter{subfigure}{0}
    \subfloat[Input synthetic image]{
    \includegraphics[width=0.15\textwidth]{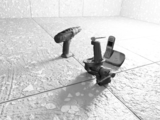}}
    \subfloat[Real image]{
    \includegraphics[width=0.15\textwidth]{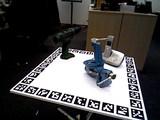}}
    \subfloat[pix2pix~\cite{isola2017image}]{
    \includegraphics[width=0.15\textwidth]{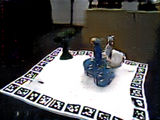}}
    \subfloat[pix2pixHD~\cite{wang2018high}]{
    \includegraphics[width=0.15\textwidth]{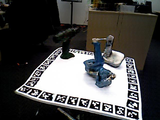}}
    \subfloat[cycleGAN~\cite{zhu2017unpaired}]{
    \includegraphics[width=0.15\textwidth]{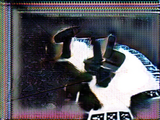}}
    \subfloat[CUT~\cite{park2020contrastive}]{
    \includegraphics[width=0.15\textwidth]{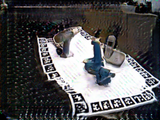}}
    \caption{\textbf{Qualitative results on HB2.} We show the results of the paired and upaired image translation baselines on the HB5 scene used in the HB-LM Cross Domain Adaptation Benchmark. Top row: images from the $HB_2$ training set; last two rows: images from the $HB_2$ test set. The real image is used as pair target when training ~\cite{isola2017image,wang2018high} while~\cite{zhu2017unpaired,park2020contrastive} use an unordered set of real images as targets.}
    \label{fig:baselines_qualitative_HB_cross_domain}
\end{figure*}

\begin{figure*}
\captionsetup[subfigure]{labelformat=empty,font=scriptsize}
    \centering
    \subfloat{
    \includegraphics[width=0.15\textwidth]{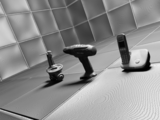}}
    \subfloat{
    \includegraphics[width=0.15\textwidth]{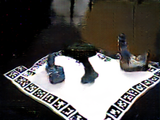}}
    \subfloat{
    \includegraphics[width=0.15\textwidth]{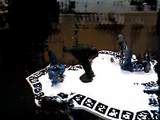}}
    \subfloat{
    \includegraphics[width=0.15\textwidth]{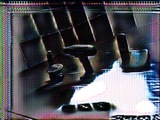}}
    \subfloat{
    \includegraphics[width=0.15\textwidth]{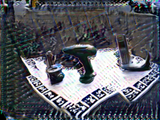}}
    \\ 
    \subfloat{
    \includegraphics[width=0.15\textwidth]{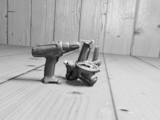}}
    \subfloat{
    \includegraphics[width=0.15\textwidth]{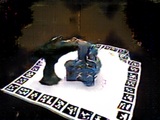}}
    \subfloat{
    \includegraphics[width=0.15\textwidth]{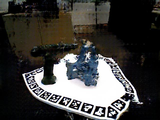}}
    \subfloat{
    \includegraphics[width=0.15\textwidth]{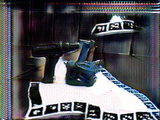}}
    \subfloat{
    \includegraphics[width=0.15\textwidth]{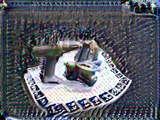}}
    \setcounter{subfigure}{0}
    \subfloat[Input synthetic image]{
    \includegraphics[width=0.15\textwidth]{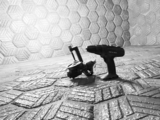}}
    \subfloat[pix2pix~\cite{isola2017image}]{
    \includegraphics[width=0.15\textwidth]{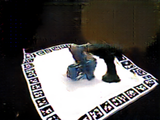}}
    \subfloat[pix2pixHD~\cite{wang2018high}]{
    \includegraphics[width=0.15\textwidth]{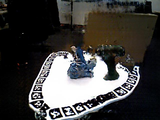}}
    \subfloat[cycleGAN~\cite{zhu2017unpaired}]{
    \includegraphics[width=0.15\textwidth]{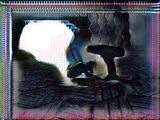}}
    \subfloat[CUT~\cite{park2020contrastive}]{
    \includegraphics[width=0.15\textwidth]{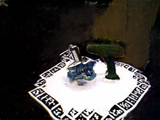}}
    \caption{\textbf{Qualitative results on synthetic images containing the HB2 objects rendered with random poses.} This is a completely synthetic dataset (i.e., no real image counterparts) and all GAN baseline methods struggle to generalize to generalize to this setting.}
    \label{fig:baselines_qualitative_random_poses}
\end{figure*}

Recall the two benchmarks \textit{HB Dynamic Lighting Benchmark} where we train on $HB_5$ and \textit{HB-LM Cross Domain Adaptation Benchmark} where we train on $HB_2$. For both benchmarks the training set consists of 272 frames, from which we generate an extended training set of 1088 with \textit{BlenderProc}~\cite{denninger2019blenderproc} using the posed CAD models and with randomized materials, as described in the main text. Using the synthetically generated images along with the corresponding real images (paired or unpaired), we train the GAN baselines as described below. A preprocessing step involves converting the synthetic images to grayscale, to avoid the complication of having the same synthetic object with multiple random materials being mapped to the same real object. Qualitative results for the \textit{HB Dynamic Lighting Benchmark} are shown in Fig.~\ref{fig:baselines_qualitative_HB_lighting} while Fig.~\ref{fig:baselines_qualitative_HB_cross_domain} shows qualitative results for the \textit{HB-LM Cross Domain Adaptation Benchmark}. Additionally, in Fig.~\ref{fig:baselines_qualitative_random_poses} we show qualitative results when running the baselines on the additional 1000 frames generated with BlenderProc containing randomized camera viewpoints and the same objects as $HB_2$ but with randomized poses. 

% For each benchmark, the GAN baselines are trained on the extended training set of 1088 images that corresponds to that benchmark, as described below.


\subsection{Paired image translation}
\label{subsec:supp_paired_baselines}

We compare PNDR against two paired image translation GAN~\cite{goodfellow2020generative} based baselines: pix2pix~\cite{isola2017image} and pix2pixHD~\cite{wang2018high}. To train these methods we use synthetic and real image pairs. Although this setting is unrealistic in practice, we leverage this information from the HB dataset and train paired image-to-image translation, which we regard as an upper bound for in-domain performance. 

\noindent\textbf{pix2pix~\cite{isola2017image}}: we use the official implementation~\cite{isola2017image,zhu2017unpaired} and train for 200 epochs with the Adam optimizer~\cite{kingma2014adam} and with a starting learning rate of $2e^{-4}$ and with $\beta_1=0.5$ and $\beta_2=0.999$. The learning rate is kept fixed for the first 100 epochs and decayed to $0$ during the next 100 epochs. The input images are resized to 286$\times$286 and a random crop of 256$\times$256 pixels is selected.

\noindent\textbf{pix2pixHD~\cite{wang2018high}}: we use the official implementation and train for 200 epochs the Adam optimizer~\cite{kingma2014adam} and with a starting learning rate of $2e^{-4}$ and with $\beta_1=0.5$ and $\beta_2=0.999$. The learning rate is kept fixed for the first 100 epochs and decayed to $0$ during the next 100 epochs. We train on the original image resolution of 640$\times$480 without any resizing and without any cropping.

\subsection{Unpaired image translation}
\label{subsec:supp_unpaired_baselines}

We also compare PNDR against the unpaired image translation baselines cycleGAN~\cite{zhu2017unpaired} and CUT~\cite{park2020contrastive}. As before, we first convert the synthetic images to grayscale to simplify learning. Nevertheless, we empirically observed poor translation results, and particularly the GANs failed to maintain the shape of the objects, which is crucial to our downstream task, i.e., correspondence-based 3D object detection. To improve the performance of these baselines, we leverage the ground truth object masks, and added an L1 loss between the input grayscale synthetic image and the image outputted by the generator, which we first convert to grayscale as well. The L1 loss is applied only on the pixels falling in the ground truth object mask. Although unrealistic in practice, this additional L1 loss helped the unpaired image translation method maintain the shape of the objects in the generated images.  

\noindent\textbf{cycleGAN~\cite{zhu2017unpaired}}: we use the official implementation ~\cite{isola2017image,zhu2017unpaired} and train for 200 epochs with the Adam optimizer~\cite{kingma2014adam} and with a starting learning rate of $2e^{-4}$ and with $\beta_1=0.5$ and $\beta_2=0.999$. The learning rate decay and input image resolution are the same as when training \textit{pix2pix}, the only difference being the L1 loss between the generated image and the input synthetic image.

\noindent\textbf{CUT~\cite{park2020contrastive}}: we use the official implementation~\cite{park2020contrastive} and train for 400 epochs with the Adam optimizer~\cite{kingma2014adam} with $\beta_1=0.5$ and $\beta_2=0.999$. The starting learning is $2e^{-4}$ and is kept fixed for 200 epochs after which it is decayed linearly to 0 over the next 200 epochs. The input images are resized to 286$\times$286 and a random crop of size 256$\times$256 is selected during training. We apply the same L1 loss between the generated output image and the input synthetic image on the pixels contained in the ground truth object mask.

\subsection{Monocular Depth Estimation}
\label{sec:monodepth}

% \subsubsection{Implementation details.}

\noindent\textbf{monodepth2~\cite{godard2019digging}}: We use the ResNet18~\cite{he2016deep} encoder-decoder architecture, as described in~\cite{godard2019digging}, \textit{without} ImageNet pretraining. We train for 20 epochs with the Adam optimizer~\cite{kingma2014adam} with 
$\beta_1=0.9$ and $\beta_2=0.999$. The learning rate starts at $1.5e^{-4}$ as is decayed by 20\% every 5 epochs and the input images are resized to 256 $\times$ 320. We use standard color jittering and no cropping.

\noindent\textbf{packnet-sfm~\cite{guizilini20203d}}: we use the PackNet architecture from the official implementation. We train for 20 epochs with the Adam optimizer~\cite{kingma2014adam} with 
$\beta_1=0.9$ and $\beta_2=0.999$. The learning rate starts at $2e^{-4}$ as is decayed by 20\% every 5 epochs and the input images are resized to 256 $\times$ 320. We use standard color jittering and no cropping.

\subsubsection{Losses.}

We follow the standard monocular depth estimation losses~\cite{eigen2014depth} defined as follows:
\begin{equation}
AbsRel=\frac{1}{N} \sum_{d \in D^*} \frac{|d - d^*|}{d^*} 
\end{equation}

\begin{equation}
RMSE=\sqrt{\frac{1}{|D^*|} \sum_{d \in D^*} |d - d^*|}
\end{equation}

\begin{equation}
\delta_1 =\text{\% of d s.t. } \max\left(\frac{d}{d^*}, \frac{d^*}{d}\right) < 1.25
\end{equation}

where $d^*$ and $d$ represents respectively ground-truth and corresponding predicted depth values, with $D^*$ being the set containing all valid ground-truth depth pixels.

\setlength{\tabcolsep}{20pt}
\begin{table}[b]
  \centering
  \caption{\textbf{2D detection and instance segmentation}.}
  \resizebox{1\columnwidth}{!}{
    \begin{tabular}{c|c|cc|cc}
    \toprule
    \multirow{2}[4]{*}{\textbf{Train}} & \multirow{2}[4]{*}{\textbf{Method}} & \multicolumn{2}{c|}{\textbf{HB5}} & \multicolumn{2}{c}{\textbf{HB10}} \\
\cmidrule{3-6}          &       & \textbf{F1} & \textbf{mIoU} & \textbf{F1} & \textbf{mIoU} \\
    \midrule
    Real  &       & 0.92  & 0.95  & 0.21  & 0.46 \\
    \midrule
    \multicolumn{1}{c|}{\multirow{2}[2]{*}{CAD}} & RayTraced - 1088 & 0.54  & 0.74  & 0.07  & 0.34 \\
          & PNDR - 1088  & 0.61  & 0.83  & 0.22  & 0.51 \\
    \bottomrule
    \end{tabular}%
    }
  \label{tab:evaluation}%
\end{table}%

\subsection{Other Downstream Tasks} 
We extend the CBOD evaluation to support these tasks on the HB dynamic lighting benchmark. In particular, we use CBOD's multi-label object masks to compute 2D bounding boxes. We use a standard F1 metric, which is defined as a weighted average of precision and recall, to evaluate performance of 2D object detection. We consider detections to be correct when the intersection over union (IoU) between predicted and ground truth bounding boxes is $>=0.5$. We also evaluate the quality of estimated multi-label object masks using the IoU metric. Our results on the new tasks are summarized in Table~\ref{tab:evaluation} and they are consistent with our observations on the tasks of 6D object detection and monocular depth estimation (cf. Tables 1, 2, and 5 in the main paper). For a constant light scene (i.e. HB5), PNDR improves over training on domain-randomized ray tracing images, closing the gap towards real data training. Additionally, PNDR achieves much better generalization to a more difficult dynamic lighting scene (i.e. HB10), improving even over the baseline trained directly on real data. 
\end{document}